\documentclass[runningheads]{llncs}
\usepackage{graphicx}
\usepackage{amsmath}
\usepackage[hidelinks]{hyperref}

\usepackage{tabularx}

\renewcommand\vec{\mathbf}

\begin{document}
\title{Traffic Camera Calibration via Vehicle Vanishing Point Detection}
\author{Viktor Kocur\inst{1}\orcidID{0000-0001-8752-2685} \and
Milan Ft\'{a}\v{c}nik\inst{1}}
\authorrunning{V. Kocur and M. Ft\'{a}\v{c}nik}
\institute{Faculty of Mathematics, Physics
and Informatics of Comenius University in Bratislava, Mlynská Dolina, 841 01, Bratislava, Slovakia}
%\and
%Springer Heidelberg, Tiergartenstr. 17, 69121 Heidelberg, Germany
\maketitle              % typeset the header of the contribution
\begin{abstract}

In this paper we propose a traffic surveillance camera calibration method based on detection of pairs of vanishing points associated with vehicles in the traffic surveillance footage. To detect the vanishing points we propose a CNN which outputs heatmaps in which the positions of vanishing points are represented using the diamond space parametrization which enables us to detect vanishing points from the whole infinite projective space. From the detected pairs of vanishing points for multiple vehicles in a scene we establish the scene geometry by estimating the focal length of the camera and the orientation of the road plane. We show that our method achieves competitive results on the BrnoCarPark dataset while having fewer requirements than the current state of the art approach.

\end{abstract}
\section{Introduction}

Automatic traffic surveillance aims to provide information about the surveilled vehicles such as their speed, type and dimensions and as such is an important aspect of intelligent transportation system design. To perform these tasks an automatic traffic surveillance system requires an accurate calibration of the recording equipment. 

Standard procedures of camera calibration require a calibration pattern or measurement of distances on the road plane which requires human intervention. Ideally the camera calibration should be carried out automatically. Traffic cameras can be automatically calibrated when observing straight road segments \cite{dubska2014,sochor2017}. The limitation to straight road segments makes the methods not applicable in various important traffic scenarios such as curved roads, intersections, roundabouts and parking lots. Methods not requiring straight road segments also exist, but they come with specific drawbacks such as requiring parallel curves to be visible on the road plane \cite{parallelcurves} or vehicles of specific make and model to be present in the scene \cite{planecalib,landmarkcalib,autocalib}. 

In this paper we propose a camera calibration method which overcomes these limitations.\footnote{We make the code available online: \url{https://github.com/kocurvik/deep\_vp/}} Our method works by estimating pairs of orthogonal vanishing points for individual vehicles present in the scene. For this purpose we use a deep convolutional neural network based on the stacked hourglass architecture \cite{hourglass}. The outputs of the network are interpreted as heatmaps of the poistions of vanishing points represented in the diamond space parametrization \cite{diamondspace}. The diamond space parametrization is used as it provides a convenient way of representing the whole infinite projective space in a bounded diamond shaped space. To minimize the error caused by parametrizing an infinite space by a heatmap with finite resolution we opt to output multiple heatmaps for each of the vanishing points with each heatmap representing the original projective space at a different scales. During inference we select the heatmap with the lowest error caused by the sampling of the grid around the maximum of the peak of the heatmap. Finally, the obtained pairs of vanishing points for vehicles across multiple frames are used to determine the horizon line of the road plane and the focal length of the camera in a simple manner inspired by the Theil-Sen estimator \cite{theil}.

We evaluate our method on two different datasets of traffic surveillance videos of straight roads \cite{brnocompspeed} and parking lots \cite{landmarkcalib}. Our experiments show that our method is not competitive against more specialized methods which utilize the assumption of a straight road segment. However when considering the parking lot scenario where these methods are not applicable and our method achieves accuracy close to the existing state of the art approach \cite{landmarkcalib} while not being limited by requiring vehicles of specific make and model to be present in the scene.

We also experimentally show that approaching this problem as a direct regression task of finding the two vanishing points leads to poor accuracy or even an inability for the network to converge depending on the used loss function. Additionally we also verify the impact of using multiple heatmaps instead of one for each vanishing point as well as the effect of training data augmentation.

\section{Related work}

\subsection{Traffic camera calibration}

In the context of traffic surveillance the knowledge of the scene geometry can be used to infer the real-world positions of the surveilled vehicles. In order to obtain the geometry of the scene it is necessary to perform camera calibration. A review of available methods has been presented by Sochor et al. \cite{brnocompspeed}. The review found that most published methods are not automatic and require human input such as drawing a calibration pattern on the road, using positions of line markings on the road or some other measured distances related to the scene.

%In the context of traffic surveillance the knowledge of the scene geometry can be used to infer the real-world positions of the surveilled vehicles. In order to obtain the geometry of the scene it is necessary to perform camera calibration. A review of available methods has been presented by Sochor et al. \cite{brnocompspeed}. The review found that most published methods are not automatic and require human input such as drawing a calibration pattern on the road \cite{pattern}, using positions of line markings on the road \cite{cathey,lan,maduro} or some other measured distances related to the scene \cite{schoepflin,you}.

An automated method based on the detection of two orthogonal vanishing points has been proposed by Dubská et al. \cite{dubska2014}. The first vanishing point corresponds to the movement of the vehicles. Relevant keypoints are detected and tracked using the KLT tracker. The resulting lines connecting the tracked points between frames are then accumulated using the diamond space accumulator \cite{diamondspace}. The second vanishing point is accumulated from the edges of moving vehicles which are not aligned with the vehicle motion. The two orthogonal vanishing points are sufficient to compute the orientation of the road plane relative to the camera and the focal length. In order to enable the measurement of real-world distances in the road plane a scale parameter is determined by constructing 3D bounding boxes around detected vehicles and recording their mean dimensions which are then compared to statistical data based on typical composition of traffic in the country. This method has been further improved \cite{sochor2017} by fitting a 3D model of a known common vehicle to its detection in the footage. The detection of the second vanishing point is also improved by using edgelets instead of edges. These methods are fully automatic and provide accurate camera calibration, however they only work when the scene contains a straight segment of a road thus preventing their use in many important traffic surveillance scenarios such as intersections, roundabouts, curved roads or parking lots.

Corral-Soto and Elder \cite{parallelcurves} detect pairs of curves which are parallel in the scene and use the properties of parallel curves under perspective transformation to iteratively pair points on one curve with their counterparts on the other curve followed by the optimization of the camera parameters. This approach works for both straight as well as curved roads, but requires the presence of clear parallel curves in the scene which may not be present in many traffic surveillance scenarios. 

More recent methods rely on detecting landmark keypoints of specific vehicle models. The relative 3D positions of these keypoints are known for each vehicle and therefore they can be used to obtain the extrinsic camera parameters by using a standard PnP solver. Bhardwaj et al. \cite{autocalib} use extensive filtering and averaging to obtain the final extrinsic parameters. Bartl et al. \cite{planecalib} use iterative calculation of the focal length to minimize the weighted reprojection error of the keypoints followed by a least squares estimation of the ground plane. In a very similar approach \cite{landmarkcalib} the authors employ differential evolution to obtain all of the camera parameters in single optimization scheme. These methods require that few specific models of vehicles are present in the scene, which may prevent their use in regions with different vehicle model composition or in areas with low traffic density. Only the last two methods do not require the focal length to be known therefore we use them as benchmarks for evaluation of our method.

\subsection{Vanishing point detection}

Since vanishing points provide important information about scene geometry a range of methods of detecting the vanishing points have been proposed. Methods usually rely on various estimation algorithms such as the Hough transform \cite{vphough}, EM \cite{vpem}, RANSAC \cite{vpransac} or deep neural networks \cite{deepvp}. 

In our approach we use a deep neural network to detect the vanishing points parametrized in the diamond space which was introduced in \cite{diamondspace} where it was used in a modified scheme of the Cascaded Hough Transform \cite{CHT} to obtain three orthogonal vanishing points of a Manhattan world. The diamond space parametrization is based on the PClines line parameterization \cite{pclines} and it is defined as a mapping of the whole real projective plane to a finite diamond shaped space. We discuss the properties of this mapping in greater detail in section \ref{sec:diamond}.

%\subsection{Stacked hourglass networks}

%The stacked hourglass architecture \cite{hourglass} was first introduced for the task of human keypoint detection. The architecture relies on the successive steps of pooling and upsampling that are done to produce a final set of predictions heatmaps. The repeated bottom-up, top-down processing used in conjunction with intermediate supervision enables the network to capture the various spatial relationships useful for the task of keypoint detection.

%The architecture has also been sucessfully adapted for the task of object detection \cite{CornerNet}. This development inspired further research into state of the art keypoint-based object detection \cite{TripleCenterNet,CenterNet}. 

\section{Traffic camera geometry}

Under perspective projection the images of parallel lines meet in the vanishing point corresponding to the direction of the lines. The knowledge of vanishing points can be utilized to constrain the camera calibration parameters. Each pair of vanishing points corresponding to two orthogonal directions places a single constraint on the focal length and the coordinates of the principal point. Therefore knowledge of three orthogonal vanishing points is sufficient to obtain the intrinsic camera parameters when no distortion and zero skew is assumed. If the position of the principal point is known or can be reliably assumed then only one such pair is sufficient. Additionally, the rotation of the camera with respect to the directions represented by the vanishing points can also be obtained and used to project image points onto the plane parallel with these directions.

Given the pripcipal point $\vec{p} = (p_x, p_y)$ and two orthogonal vanishing points $\vec{u} = (u_x, u_y)$ and $\vec{v} = (v_x, v_y)$ all expressed in the pixel coordinates and assuming zero skew and no distortion is is possible to calculate the focal length of the camera:
\begin{equation}
f = \sqrt{-(\vec{u} - \vec{p}) \cdot (\vec{v} - \vec{p})} = \sqrt{- (u_x - p_x) (u_y - p_y) - (v_x - p_x) (v_y - p_y)}\,.
\label{eqn:focal}
\end{equation}
Note that for some configurations this leads to imaginary focal lengths which are not physically meaningful. In practice we discard any such configurations as invalid. 

The two vanishing points also define a line $\vec{h} = (h_a, h_b, h_c)$ in the image known as the horizon line. The horizon line corresponds to a set of planes which are parallel with the directions corresponding to the two vanishing points. The shared normal of these planes can then be calculated using the intrinsic matrix $K$:

\begin{equation}
\vec{n} = K^T \vec{h} = 
\begin{pmatrix}
f & 0 & 0  \\
0 & f & 0 \\
p_x & p_y & 1
\end{pmatrix}    
\begin{pmatrix}
h_a \\
h_b \\
h_c
\end{pmatrix}\,.
\label{eqn:plane_normal}
\end{equation}

It is possible to use the normal $\vec{n}$ to project any point $\vec{q} = (q_x, q_y)$  in the image to a point $\vec{Q}$ lying in a plane defined by the normal:

\begin{equation}
\vec{Q} = -\frac{\delta}{\vec{\hat{q}} \cdot \vec{n}} \vec{\hat{q}}\,,
\label{eqn:proj}
\end{equation} 
where $\vec{\hat{q}}$ denotes the vector $(q_x - p_x, q_y - p_y, f)$. Note that due to the ambiguity arising from the properties of single-view perspective geometry only the normal of the plane is established. Therefore in order for the projected points to have accurate metric properties it is necessary to include a scaling parameter $\delta$, which needs to be determined for the plane by using a real world measurement. If $\delta$ is not established it is still possible to use this projection with an arbitrary non-zero $\delta$ to obtain relative distances of points on the plane of interest.

Since we want to detect two vanishing points for every vehicle we establish the directions to which these two would correspond to. In further text we will denote the vanishing point corresponding to the direction in which the vehicle is facing as the \textit{first vanishing point}. The \textit{second vanishing point} corresponds to the direction parallel with the road plane and orthogonal to the direction in which the vehicle is facing. This direction is also parallel to the vehicle axles. 

\section{Diamond Space}

\label{sec:diamond}

To represent the vanishing points we use the diamond space parametrization \cite{diamondspace}. This parametrization can be thought of as a mapping of the infinite projective space to a bounded diamond-shaped space. The relation (\ref{eqn:to_diamond}) shows the mapping of a point expressed in homogeneous coordinates from the original projective space to the diamond space. The relation (\ref{eqn:to_original}) shows the opposite mapping.

\begin{equation}
(x, y, w)_o \rightarrow (-w, -x, \textnormal{sgn}(xy)x + y + \textnormal{sgn}(y)w)_d
\label{eqn:to_diamond}
\end{equation}

\begin{equation}
(x, y, w)_d \rightarrow (y, \textnormal{sgn}(x)x + \textnormal{sgn}(y)y - w, x)_o
\label{eqn:to_original}
\end{equation}

\begin{figure}[t]
    \centering
    \includegraphics[width=\textwidth]{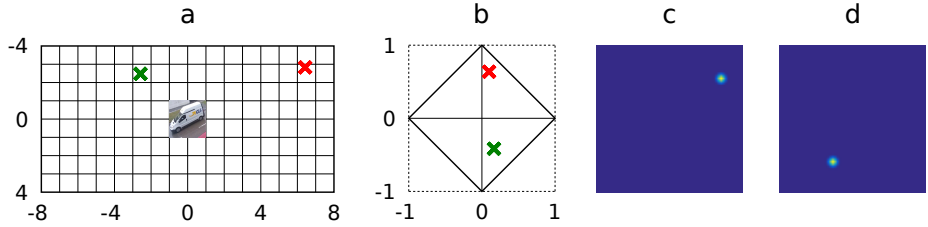}
    \caption{a) The first (red cross) and the second (green cross) vanishing point in an image coordinate system where the dimensions of the bounding box of a vehicle span from -1 to 1 for both axes. b) The same vanishing points represented in the diamond space. c) The heatmap generated for the first vanishing point. d) The heatmap generated for the second vanishing point. }
    \label{fig:diamond}
\end{figure}

In our method we represent the diamond space with a heatmap of finite resolution. For use in the heatmap we also rotate the diamond space by 45 degrees clockwise for a more efficient representation. The ground truth vanishing points are represented as 2D Gaussians with standard deviation of 1 pixel centered on the pixel closest to the actual position of the vanishing point in the diamond space. An example of the vanishing points in the original image space, the diamond space and the heatmaps is shown in Fig. \ref{fig:diamond}.

Representing the diamond space with a heatmap of finite resolution brings about an issue of insufficient representation for some positions of vanishing points. In general the points which are farther from the origin in the original image space will have a greater systemic error arising from insufficient heatmap representation. To reduce the error the original image space can be scaled so that the vanishing points are closer to the origin. However just choosing only one scale may lead to problems since then the objects which were close to the origin before the scaling would occupy smaller space in the image space and thus also be represented with greater error in the heatmap. Therefore we choose to use multiple scales to represent each vanishing point. During inference we select the value from the heatmap with the smallest expected error.

\section{Vanishing point detection network}

To detect the vanishing points we use the stacked hourglass network \cite{hourglass} with two hourglass modules. We make a slight modification in the initial convolutional layer of the network by not using stride. As input we use cropped images containing only one vehicle with the resolution $128 \times 128$ pixels. The network outputs 4 heatmaps with the resolution $64 \times 64$ representing the diamond space at four different scales (0.03, 0.1, 0.3, 1.0) for each of the two vanishing points.

For training we use the data from the BoxCars116k dataset \cite{boxcars}. The dataset contains images of vehicles which were cropped from traffic surveillance footage along with bounding box and vanishing point annotations. We split the dataset into 92 932 training images, 11 798 validation images and 11 556 testing images.

We train the network with L2 loss over the heatmaps with batch size of 32 using the Adam optimizer. We train with the learning rate at 0.001 for 60 epochs and 0.0001 for 15 epochs.

\subsection{Augmentation}

\begin{figure}[t]
    \centering
    \includegraphics[width=\textwidth]{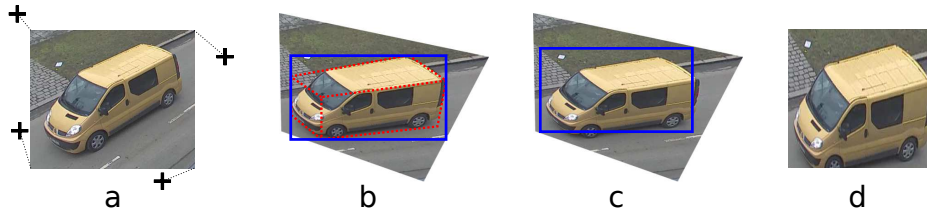}
    \caption{Training data augmentation. a) The original image from the BoxCars116k dataset \cite{boxcars}. Each corner (black cross) is randomly perturbed to generate a perspective transformation. b) The image after perspective transformation. The transformation is also applied to the ground truth 3D bounding box (dotted red). 2D bounding box (solid blue) is constructed based on the 3D bounding box. c) The 2D bounding box is randomly perturbed. d) The image is cropped to only contain the contents of the perturbed 2D bounding box and rescaled to $128 \times 128$ pixels.}
    \label{fig:aug}
\end{figure}

The BoxCars116k dataset contains only images of vehicles captured by 137 distinct traffic cameras. Therefore the distribution of vanishing point positions in the training data is limited. To remedy this problem we apply a random perspective transformation on the training images and vanishing points.

The perspective transform is constructed by warping the position of the corners of the original image by a random offset sampled from normal distribution with a standard deviation of 12.5 pixels. The transformation is applied on the image, the two vanishing points and the ground truth 3D bounding box. The transformed 3D bounding box is used to calculate the new 2D bounding box of the vehicle. Since vehicles are not planar objects this approach is not strictly geometrically correct, but it nevertheless results in better performance.

During inference the network will work in a pipeline after detecting vehicles with an independent object detection network. Since the object detection network may output badly aligned bounding boxes we also augment the 2D bounding box by shifting each corner with values sampled from uniform distribution spanning -5 to 5 pixels. We also horizontally flip the images with a 50 \% probability. The whole process of augmentation of a single training sample is visualised in Fig \ref{fig:aug}.

\begin{figure}[t]
    \centering
    \includegraphics[width=\textwidth]{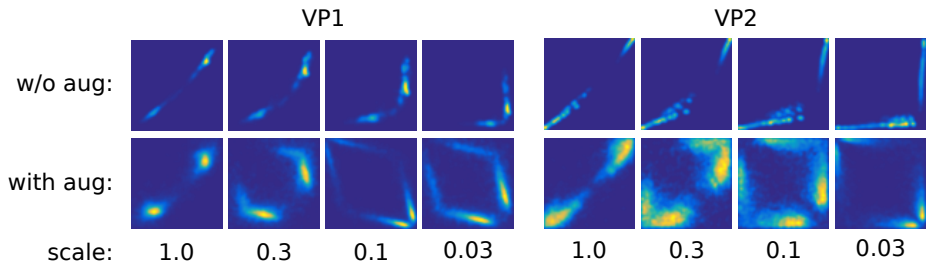}
    \caption{The cummulative heatmaps for the first (left) and the second (right) vanishing point at the four scales used by the network averaged over 10 000 training samples without augmentation (top) and with augmentation (bottom). Without augmentation most of the peaks are concentrated in a small part of the heatmap. After applying augmentation the training data represent a much broader distribution of vanishing point positions. }
    \label{fig:cummulative_heatmap}
\end{figure}

Applying the transformation significantly expands the range of vanishing point positions in the training data. This fact is visualised in Fig. \ref{fig:cummulative_heatmap}.

\section{Camera calibration pipeline}

\subsection{Vanishing point detection}

To calibrate a traffic camera we receive its traffic surveillance footage and detect vehicles in the individual frames using the CenterNet object detector \cite{CenterNet} pre-trained on the MS COCO dataset \cite{MSCOCO}. We only retain at most ten bounding boxes with greatest confidence. In scenes with vehicles which are static over multiple frames we discard the static vehicles after few initial detections. With the remaining bounding boxes we crop the detected vehicles from the frames and resize them to $128 \times 128$ pixels. These vehicle images are then put into the vanishing point detection network to obtain the 4 heatmaps for each vanishing point. For each scale we find the row $i^s$ and column $j^s$ with the maximum heatmap value. We also find all of the positions in the heatmap for which the value is close to the maximum:

\begin{equation}
P^s = \left\{(i, j) \mid H_{i, j}^s \ge 0.8 H_{i^s, j^s}^s \right\},
\end{equation}
where $H_{i, j}^s$ is the value in the $i$-th row and $j$-th column in the heatmap for scale $s$. We then calculate an approximate measure of accuracy of the vanishing point for each scale:

\begin{equation}
D^s = \frac{1}{|P^s|} \sum_{(i, j) \in P^s} \frac{|| \vec{v}_{i, j}^s - \vec{v}_{i^s, j^s}^s ||}{|| \vec{v}_{i^s, j^s}^s||},
\end{equation}
where $|P^s|$ is the number of elements in the set, $\vec{v}_{i, j}^s$ is the vanishing point corresponding to the $i$-th row and $j$-th column in the heatmap for scale $s$ in the image coordinate system in which the vehicle bounding box is centered at the origin and spans $-1$ to $1$ in both dimensions (see Fig. \ref{fig:diamond}-a). We then select the vanishing point $\vec{v}_{i^s, j^s}^s$ with the scale which minimizes $D^s$. The measure $D^s$ incorporates both the inaccuracies arising from the finite resolution of the heatmaps as well as the uncertainty caused by unclear peaks in the heatmaps. Finally we convert the vanishing point position from the coordinate system tied to the bounding box to the image coordinate system of the whole frame.

\subsection{Calculating the scene geometry}

Given the pairs of vanishing poitns we first calculate the focal lengths for each pair using (\ref{eqn:focal}). We discard all pairs which result in an imaginary focal length. We determine the focal length to be the median of the remaining focal lengths. Assuming zero skew, no distortion and the principal point of the camera to be in the centre of the image we now construct the intrinsic camera matrix $K$.

We determine the horizon using an approach inspired by the Theil-Sen estimator \cite{theil}. For each pair of vanishing points we determine the line defined by them and express it in the form of $y = mx + k$. We then take the median of all the values of $m$ and denote it $\hat{m}$. For each vanishing point $\vec{v} = (v_x, v_y)$ we calculate $q = v_y - v_x \hat{m}$. Again we calculate the median of the values of $q$ and denote it as $\hat{q}$. The final horizon can then be expressed with the equation $y = \hat{m} x + \hat{q}$. Converting to the form $ax + by + c = 1$ we get the line $\vec{h} = (\hat{m}, -1 ,\hat{q})$. We can then use $(\ref{eqn:plane_normal})$ to determine the normal of the road plane.

\section{Evaluation}

We evaluate our method on two traffic surveillance datasets BrnoCompSpeed \cite{brnocompspeed} and BrnoCarPark \cite{landmarkcalib}. Both datasets contain information about measured distances between pairs of points on the road plane along with their pixel positions in the images. We use this information to evaluate the accuracy of our methods. Since our method does not produce a scale we evaluate the accuracy of the camera calibration by comparing the relative differences of ratios of two different measurements defined as

\begin{equation}
r_{i, j} = \frac{\left|\frac{\hat{d}_i}{\hat{d}_j} - \frac{d_i}{d_j} \right|}{\frac{\hat{d}_i}{\hat{d}_j}},
\end{equation}
where $\hat{d}_i$ is the $i$-th ground truth distance measurement and the $d_i$ is the $i$-th measurement based on the projection (\ref{eqn:proj}) for the given method. Since this type or error does not depend on the scaling parameter $\delta$ we set it to one. To obtain the calibration error for a single camera we compute the mean over all of the possible pairs of ground truth measurements.

\begin{table}
\begin{minipage}[t]{0.45\linewidth}
\centering
\begin{tabular}{|c|c|}
\hline
Method & Mean error \\ \hline
DubskaCalib \cite{dubska2014} & 8.54 \\ \hline
SochorCalib \cite{sochor2017} & 3.83\\ \hline
DeepVPCalib (ours) & 14.95 \\ \hline
\end{tabular}
\caption{The mean calibration error on the split C of the BrnoCompSpeed dataset \cite{brnocompspeed}. \label{tab:bcs}}
\end{minipage}
\hfill
\begin{minipage}[t]{0.45\linewidth}
\centering
\begin{tabular}{|c|c|}
\hline
Method & Mean error \\ \hline
LandmarkCalib \cite{landmarkcalib} & 8.17 \\ \hline
PlaneCalib \cite{planecalib} & 21.99 \\ \hline
DeepVPCalib (ours) & 8.66 \\ \hline
\end{tabular}
\caption{The mean calibration error on the BrnoCarPark dataset \cite{landmarkcalib}. \label{tab:bcp}}
\end{minipage}
\end{table}

\subsection{BrnoCompSpeed}

The BrnoCompSpeed dataset \cite{brnocompspeed} contains 21 hour-long videos of a freely flowing traffic on straight road segments captured from overpasses. For evaluation we use the split C of the dataset which contains 9 videos. For our method we only use every tenth frame of the video and we only process the first 1500 frames.

We compare our method denoted as \textit{DeepVPCalib} with the fully automatic methods from \cite{dubska2014} which we denote as \textit{DubskaCalib} and \cite{sochor2017} which we denote as \textit{SochorCalib}. The results are available in the Table \ref{tab:bcs}. Compared with the state of the art our method underperforms. However we note that the methods \textit{DubskaCalib} and \textit{SochorCalib} both only work if the observed vehicles move along a straight road segment and our method does not have this limitation.

\subsection{BrnoCarPark}

The BrnoCarPark dataset \cite{landmarkcalib} contains footage from 11 different cameras observing parking lots. The footage is available both in video form and as frames which were directly used for camera calibration by the method which was published alongside the dataset. To perform a better comparison with this method which we denote as \textit{LandmarkCalib} we also apply our method denoted as \textit{DeepVPCalib} on the selected frames and not the whole video. The number of frames per session ranges from 152 to 45 282, but we opt to use at most the first 5000 frames. In the comparison we also include the results from \cite{planecalib} which we denote as \textit{PlaneCalib}. The results for the methods are available in the Table \ref{tab:bcp}.

The calibration error for our method is comparable to the error of the state of the art method \textit{LandmarkCalib} while not requiring any vehicles of specific models to be present in the surveilled scene. Our method is thus more easily usable in wider range of geographical regions where the composition of vehicle models might be such that any traffic surveillance footage might contain only a very limited number of the required vehicles to be visible. 

\subsection{Ablation studies}

\label{sec:ablation}

\begin{table}[]
\resizebox{\linewidth}{!}{
\begin{tabular}{|c|c|c|c|c|c|c|c|c|c|}
\cline{6-10}
\multicolumn{1}{c}{}  &  \multicolumn{1}{c}{}  &  \multicolumn{1}{c}{}  &  \multicolumn{1}{c}{}  &  \multicolumn{1}{c}{} &\multicolumn{3}{|c|}{Augmentation}   & \multicolumn{2}{c|}{Mean calibration error} \\ \hline
\# & Type & Backbone & Scales & Loss & Perspective & BBox shift & Flip & BrnoCompSpeed   & BrnoCarPark   \\ \hline
1 & heatmap &  hourglass &  0.03, 0.1, 0.3, 1.0 & L2 & 12.5 px & 5 px & yes & \textbf{14.94} & \textbf{8.66}  \\ \hline
2 & heatmap &  hourglass &  0.03, 0.1, 0.3, 1.0 & L2 & 5 px & 5 px & yes & 18.29 & 12.41  \\ \hline
3 & heatmap &  hourglass &  0.03, 0.1, 0.3, 1.0 & L2 & n/a & n/a & no & 19.17 & 19.67  \\ \hline
4 & heatmap &  hourglass &  0.03  & L2 & 12.5 px & 5 px & yes & 19.70 & 20.40  \\ \hline
5 & heatmap &  hourglass &  0.1  & L2 & 12.5 px & 5 px & yes & 18.11 & 46.83  \\ \hline
6 & regression &  resnet50 & 1.0 & nL1 & 12.5 px & 5 px & yes & 19.48 & 20.45 \\ \hline
7 & regression &  resnet50 & 1.0 & nL1 & 5 px & 5 px & yes & 15.54 & 19.64 \\ \hline
8 & regression &  resnet50 & 1.0 & nL2 & 12.5 px & 5 px & yes & 20.57 & 22.68 \\ \hline
9 & regression &  resnet50 & 1.0 & nL2 & 5 px & 5 px & yes & 15.31 & 21.60 \\ \hline
10 & regression &  hourglass & 1.0 & nL1 & 12.5 px & 5 px & yes & 18.35 & 22.62 \\ \hline
\end{tabular}
}
\caption{Results of the various ablation experiments discussed in subsection \ref{sec:ablation}.\label{tab:ablation}}
\end{table}

To verify that the various components of our method are necessary to obtain the final results we trained several models with some of the components removed. The results on the split C of the BrnoCompSpeed dataset \cite{brnocompspeed} and the BrnoCarPark dataset \cite{landmarkcalib} for all of these models are presented in Table \ref{tab:ablation}. The model for the main method presented in this article is labeled as model 1.

The results for models 1-3 show that augmentation significantly reduces the calibration error especially on the BrnoCarPark dataset. To verify that the use of four heatmaps at different scales is beneficial we
trained models 4 and 5 with only one heatmap for each vanishing point. As expected using just a single scale leads to increased error.

Since the task of detecting vanishing points is essentially a regression task we also tested an approach based on direct regression of the positions of vanishing points. We ran multiple experiments training networks with L1 or L2 loss directly applied to the outputs of the network, but we were unable to get the networks to converge to any meaningful results even with extensive grid search for optimal hyperparameters. We therefore opt use normalized lossses. The normalized L2 loss has the form of:

\begin{equation}
loss_{nL2}(\vec{v}, \vec{\hat{v}}) = \frac{|| \vec{v} - \vec{\hat{v}} ||^2}{|| \vec{\hat{v}} ||^2},    
\label{eqn:nL2}
\end{equation}
where $\vec{v}$ is the output vanishing point and $\vec{\hat{v}}$ is ground truth vanishing point. The normalized L1 loss is analogous to (\ref{eqn:nL2}). For models 6-9 we used the standard ResNet50 backbone \cite{resnet} trained for 60 epochs with learning rate of 0.001, 20 epochs with learning rate of 0.0001 and finally 20 epochs with the learning rate of 0.00001. For model 10 we used the stacked hourglass network with 2 hourglass modules outputting a 64 channel heatmap followed by global pooling and several fully connected layers and the same training schedule as the other heatmap models. The regression models achieve significantly worse results than the main heatmap model. Therefore we conclude that using the diamond space heatmap representation is beneficial for detecting the vanishing points of vehicles for the purposes of camera calibration.

\section{Conclusion}

In this paper we have presented an automatic method for camera calibration which is based on a deep neural network detecting vanishing points of vehicles present in the traffic surveillance footage. Our method achieves competitive results on the BrnoCarPark dataset \cite{landmarkcalib} while not requiring a presence of specific vehicle models in the surveilled scene. Compared to the current state of the art our method can be applied in a wider range of use-cases without the need for further training.

In the current form our method does not provide the scaling parameter which would enable metric measurements in the road plane. A single known distance in the road plane can be used to obtain this parameter. In future we aim to make this process automatic. Our method already detects the relative orientations of vehicles via vanishing points it may be possible to use existing approaches based on 3D bounding box statistics \cite{dubska2014} or scaling of 3D CAD models of vehicles \cite{sochor2017}.

\section*{Acknowledgments}

The authors would like to thank Zuzana Kukelova for her valuable comments. The authors gratefully
acknowledge the support of NVIDIA Corporation with the donation of GPUs. 

%\subsection{Object detection}

%Recent advent of convolutional neural networks had a significant impact on the task of object detection. Two stage approaches such as Faster R-CNN \cite{FasterRCNN} use a convolutional neural network to generate proposals for objects in image. In the second stage the network determines which of these proposed regions contain objects and regress the boundaries of their bounding boxes. Mask R-CNN \cite{MaskRCNN} is an extension of Faster R-CNN which allows to additionally output the masks of the detected objects.
 
%Earlier single-stage object detection approaches \cite{RetinaNet,SSD,redmon} use a structure of anchorboxes to turn the problem into a classification and regression task followed by elimination of redundant object bounding boxes. Current state of the art approaches forego the use of anchorboxes completely and rely on detecting keypoints in the image via heatmaps on the output of the network. Various types of keypoints such as bounding box corners \cite{CornerNet}, centers \cite{CenterNet} or their combination \cite{TripleCenterNet} can be detected along with additional regression outputs such as keypoint offsets or bounding box dimensions which allow for construction of the bounding boxes of detected objects.

\bibliographystyle{splncs04}
\bibliography{main}

\end{document}